\pdfoutput=1
\documentclass[11pt]{article}
\usepackage[final]{acl}
\usepackage{times}
\usepackage{latexsym}
\usepackage[T1]{fontenc}
\usepackage[utf8]{inputenc}
\usepackage{microtype}
\usepackage{inconsolata}
\usepackage{graphicx}
\usepackage{amsmath}
\usepackage{amssymb}
\usepackage{mathtools}
\DeclareMathOperator*{\argmax}{arg\,max}
\usepackage{booktabs}
\usepackage{amssymb}
\usepackage{multicol}
\usepackage{booktabs}
\usepackage{multirow}
\usepackage{colortbl}
\usepackage{xcolor}
\usepackage{algorithm}
\usepackage{algpseudocode}
\usepackage{algorithmicx}
\usepackage{array}
\usepackage{subcaption}
\usepackage{tabularx}
\usepackage{colortbl}
\usepackage{makecell}
\usepackage{placeins}
\usepackage{float}

\newcommand{\method}{DIGA}

\title{Tricking Retrievers with Influential Tokens: An Efficient Black-Box Corpus Poisoning Attack}

\author{Cheng Wang$^\dagger$, Yiwei Wang$^{||}$, Yujun Cai$^\ddagger$, , Bryan Hooi$^\dagger$, 
\\
$^\dagger$ National University of Singapore  \quad
$^{||}$  University of California, Merced \\
$^\ddagger$ University of Queensland \\
\texttt{wcheng@comp.nus.edu.sg} 
\\}

\begin{document}
\maketitle
\begin{abstract}
Retrieval-augmented generation (RAG) systems enhance large language models by incorporating external knowledge, addressing issues like outdated internal knowledge and hallucination. However, their reliance on external knowledge bases makes them vulnerable to corpus poisoning attacks, where adversarial passages can be injected to manipulate retrieval results. Existing methods for crafting such passages, such as random token replacement or training inversion models, are often slow and computationally expensive, requiring either access to retriever's gradients or large computational resources. To address these limitations, we propose Dynamic Importance-Guided Genetic Algorithm (DIGA), an efficient black-box method that leverages two key properties of retrievers: insensitivity to token order and bias towards influential tokens. By focusing on these characteristics, DIGA dynamically adjusts its genetic operations to generate effective adversarial passages with significantly reduced time and memory usage. Our experimental evaluation shows that DIGA achieves superior efficiency and scalability compared to existing methods, while maintaining comparable or better attack success rates across multiple datasets.
\end{abstract}

\section{Introduction}
Large language models (LLMs)~\citep{qwen2, GPT-4, llama3} have shown impressive performance across a wide range of natural language processing tasks, but they still suffer from significant limitations, such as hallucination~\citep{hallucination1, hallucination2} and outdated internal knowledge. To address these issues, retrieval-augmented generation (RAG)~\citep{RAG1, RAG2} systems have emerged as a promising solution by incorporating external, up-to-date knowledge into the generation process. By retrieving relevant information from large external corpora, RAG systems can enhance the accuracy and relevance of LLM-generated outputs, especially in open-domain question answering and knowledge-intensive tasks.

Despite the benefits, the reliance on external knowledge sources exposes RAG systems to potential security vulnerabilities~\citep{zeng2024good, xue2024badrag, li2024generatingbelievingmembershipinference, anderson2024dataretrievaldatabasemembership}. A particular focus has been on adversarial attacks targeting RAG systems. Initial studies~\citep{song2020adversarial, raval2020one, liu2022order} explored attacks where adversarial passages are injected into the knowledge base to influence the language model's output for specific queries. Building upon this, \citet{corpus-poisoning} introduced a more potent form of attack termed corpus poisoning, where a single adversarial passage aims to affect retrieval results across multiple queries simultaneously. Currently, two primary approaches dominate corpus poisoning attack implementations. The first, based on HotFlip~\citep{hotflip}, initializes an adversarial passage with random tokens and iteratively optimizes it using gradients from the retriever. The second method, Vec2Text~\citep{vec2text}, originally developed to study text reconstruction from embeddings, has been repurposed for corpus poisoning by inverting the embeddings of query centroids.

\begin{table}
\small
\centering
\begin{tabular}{lccc}
\toprule
 & HotFlip & Vec2Text & \textbf{DIGA (ours)} \\
\midrule
Black-box & $\times$ & $\checkmark$ & $\checkmark$ \\
No Add. Training & $\checkmark$ & $\times$ & $\checkmark$ \\
Scalability & $\times$ & $\checkmark$ & $\checkmark$ \\
\bottomrule
\end{tabular}
\caption{\textbf{Comparison of different corpus poisoning methods.} Our proposed method is black-box based, requires no additional training, and maintains high scalability.}
\label{tab:comparison}
\end{table}

However, these existing methods face significant limitations:
\textbf{(1) White-box access:} HotFlip requires access to the retriever model's gradient, limiting its applicability in real-world scenarios where such access is often restricted.
\textbf{(2) Lack of generalizability:} Vec2Text requires training a separate inversion model on a specific dataset. This limits its applicability to new domains or datasets without retraining, as the model may struggle to reconstruct embeddings for data significantly different from its training set.
\textbf{(3) High computational demands:} Both methods require substantial computational resources, making them impractical for large-scale attacks. HotFlip's iterative optimization process is time-consuming and memory-intensive. Vec2Text, while faster in generation, requires additional memory for its inversion model and is computationally expensive during the training phase. These resource constraints severely limit the scalability of both approaches.

\begin{figure*}
    \centering
    \includegraphics[width=0.72\linewidth]{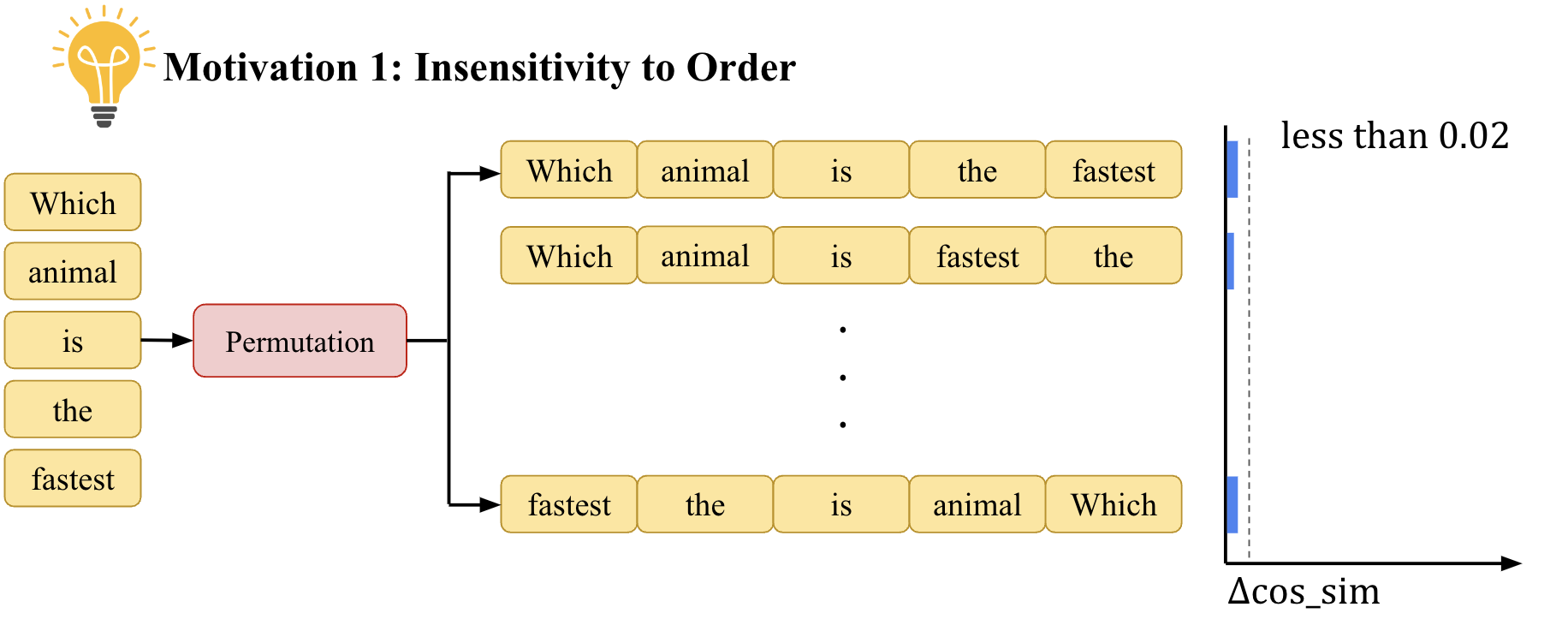}
    \includegraphics[width=0.72\linewidth]{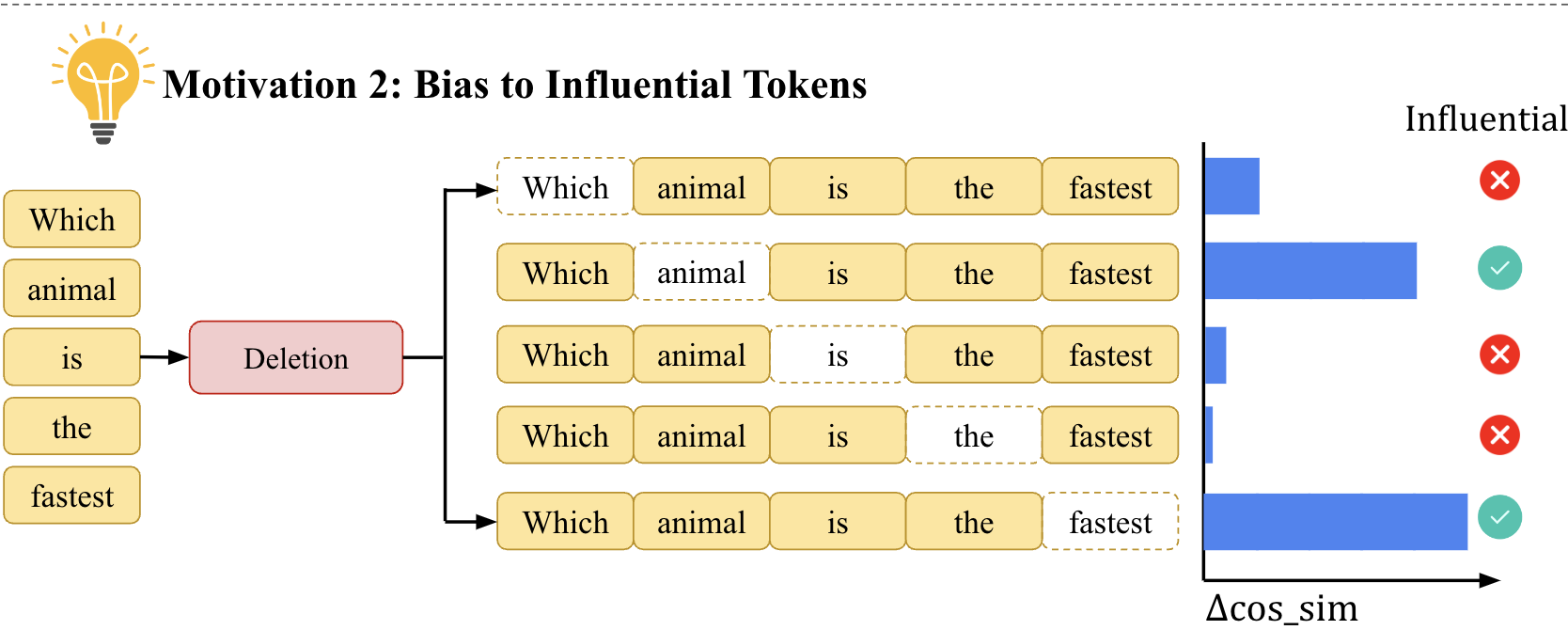}
    \caption{\textbf{Illustration of our motivations.} Top: Demonstrating insensitivity to token order, where cosine similarity remains nearly unchanged after permuting the tokens. Bottom: Highlighting bias towards influential tokens, shown by the varying effects on cosine similarity when different tokens are deleted. Some tokens are more influential since deleting them results in a larger change in similarity.}
    \label{fig:motivation}
\end{figure*}

Given the limitations of existing methods, we posit a crucial question: Is it possible to perform corpus poisoning attacks under minimal assumptions about the target system while maintaining high effectiveness and efficiency?  In response, we propose the Dynamic Importance-Guided Genetic Algorithm (\method). Our method is built upon the foundation of genetic algorithms, a class of optimization techniques inspired by the process of natural selection. \method{}~incorporates two key properties of retrieval systems into its evolutionary process:
their insensitivity to token order and bias towards influential tokens.
By leveraging these properties, \method{}~dynamically adjusts its genetic operations, such as population initialization and mutation, to craft effective adversarial passages. This approach allows us to generate potent adversarial content without requiring white-box access to the retriever or extensive computational resources (see Table~\ref{tab:comparison}). 

Our empirical evaluation demonstrates that DIGA achieves comparable or superior results to existing methods across various datasets and retrievers. Notably, DIGA accomplishes this with significantly lower time and memory usage, making it more suitable for large-scale attacks. Furthermore, our method exhibits stronger transferability across different datasets compared with another black-box method. This enhanced transferability is significant as it demonstrates DIGA's ability to generate more generalizable adversarial passages, potentially making it more effective in real-world scenarios where attack and target datasets may differ.

Our study underscores the vulnerabilities present in RAG systems and the importance of continued research into their security. Future efforts should prioritize developing robust defenses and evaluating real-world risks. We hope this work inspires further advancements in safeguarding RAG systems.

\section{Related Work}
\paragraph{Attacks on RAG Systems and Corpus Poisoning.}
Recent research has explored vulnerabilities in retrieval augmented generation (RAG) systems and dense retrievers. Membership inference attacks \citep{anderson2024dataretrievaldatabasemembership, li2024generatingbelievingmembershipinference} aim to determine the presence of specific data in knowledge bases. Other studies focus on manipulating system outputs, such as PoisonedRAG \citep{poisonedRAG} crafting misinformation-laden passages, and indirect prompt injection attacks \citep{greshake2023not} inserting malicious instructions. In corpus poisoning, \citet{corpus-poisoning} proposed a gradient-based approach inspired by HotFlip \citep{hotflip}, while Vec2Text \citep{vec2text} introduced text embedding inversion. Unlike PoisonedRAG, which targets individual queries, our work aims to craft adversarial passages affecting multiple queries simultaneously, presenting unique challenges in corpus poisoning attacks on RAG systems.
 
\paragraph{Genetic Algorithms in NLP.}
Genetic algorithms (GA) have been increasingly applied to various Natural Language Processing (NLP) tasks. Recent studies have demonstrated GA's efficacy in neural architecture search \citep{chen2024evoprompting}, adversarial prompt generation for LLM security \citep{autodan, semanticmirrorjailbreakgenetic}, and enhancing GA operations using LLMs \citep{guo2023connecting, lehman2023evolution, meyerson2023language}. These applications showcase the versatility of GA in addressing complex NLP challenges. GARAG~\citep{GARAG} shares similarities with our work in applying GA to RAG systems. However, our approach differs significantly by targeting multiple queries simultaneously, rather than focusing on query-specific adversarial passage generation. Furthermore, our method employs a dynamic mechanism specifically tailored to exploit the properties of dense retrievers, offering a more nuanced approach to the challenge of corpus poisoning in RAG systems.

\section{Method}

\subsection{Problem Formulation}
\label{sec:formulation}
Let $\mathcal{C}$ be the corpus or knowledge base of the RAG system, $Q = \{q_1, q_2, ..., q_{n}\}$ be the set of queries, $\phi_Q$ be the query embedding model, and $\phi_C$ be the context embedding model. Our goal is to generate an adversarial passage $a$ whose embedding maximizes the similarity to all the query embeddings, increasing the likelihood of retrieval for any given query. Formally, we aim to solve:

$$
a = \argmax_{a'} \frac{1}{|Q|} \sum_{q_i \in Q} \phi_Q(q_i)^T\phi_C(a').
$$

\begin{figure*}
    \centering
    \includegraphics[width=1.0\linewidth]{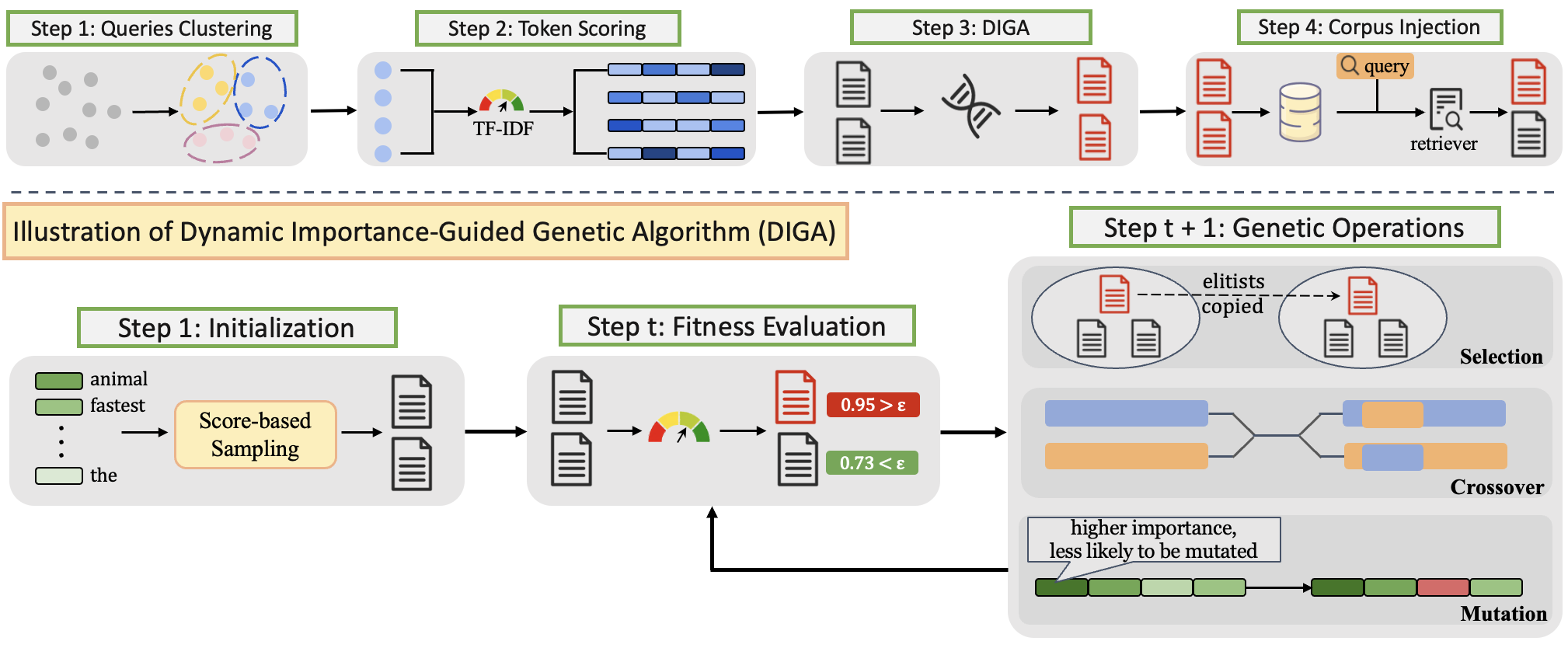}
    \caption{\textbf{An overview of our proposed method}.}
    \label{fig:pipeline}
\end{figure*}

This can be rewritten as:
\begin{align*}
    a &= \argmax_{a'} \phi_C(a')^T \frac{1}{|Q|} \sum_{q_i \in Q} \phi_Q(q_i) \\
    &= \argmax_{a'} \phi_C(a')^T \bar{\phi}_Q,
\end{align*}
where $\bar{\phi}_Q = \frac{1}{|Q|} \sum_{q_i \in Q} \phi_Q(q_i)$ represents the centroid of all query embeddings. Therefore, the adversarial passage is the solution to this optimization problem, whose embedding should maximize the similarity to the centroid of query embeddings. To generate multiple adversarial passages, we follow \citet{corpus-poisoning} by applying $k$-means clustering to the query embeddings and solving the optimization problem for each cluster centroid, resulting in $k$ adversarial passages.

\subsection{Motivation}
\label{sec:motivation}
The design of our method stems from two key observations about retrievers, as demonstrated in Figure~\ref{fig:motivation}: \textbf{(1) Insensitivity to token order:} Retrievers show remarkable insensitivity to the order of words. By comparing the original sentence with various permutations of its tokens, including extreme cases like complete reversal, we observe only negligible differences in similarity. \textbf{(2) Bias towards influential tokens:} Different words contribute unequally to the overall similarity score. When deleting certain words, the similarity drops significantly more than for others. For instance, removing "animal" or "fastest" from the sentence "Which animal is the fastest?" causes a substantial decrease in similarity, while deleting "is" or "the" has only a minor impact.

Our method leverages these properties to efficiently craft adversarial passages, resulting in a more effective and computationally efficient approach.

\section{Dynamic Importance-Guided Genetic Algorithm (DIGA)}
We propose Dynamic Importance-Guided Genetic Algorithm (DIGA), which is designed to efficiently craft adversarial passages by exploiting two key properties of retrievers: their insensitivity to token order and bias towards influential tokens. DIGA employs a genetic algorithm framework, dynamically adjusting its genetic policies based on token importance. Figure~\ref{fig:pipeline} illustrates the pipeline of our method, which consists of several key components. We will now give a detailed explanation of these components in our proposed method. We provide an algorithmic description of DIGA in Algorithm~\ref{alg:diga}. 

\subsection{Token Importance Calculation}

Our second observation—that different tokens contribute unequally to the similarity score—motivates us to quantify token importance. This quantification is crucial for guiding the genetic algorithm toward more effective adversarial passages, leading to faster convergence.

To efficiently calculate token importance across a large corpus, we employ Term Frequency-Inverse Document Frequency (TF-IDF)~\citep{TF-IDF}. Our experiments demonstrate that TF-IDF values align well with the Leave-one-out values~\citep{loo}, as illustrated in Figure~\ref{fig:motivation}. Importantly, TF-IDF calculation is considerably more efficient than LOO, making it suitable for large corpus.


\subsection{Population Initialization}
\label{sec:initialization}

Initialization is a critical component in genetic algorithms, influencing both the quality and diversity of the initial population. Our method employs a score-based initialization strategy that aligns with our observation of retrievers' bias towards influential tokens. We initialize the population by sampling tokens based on importance scores derived from TF-IDF values, ensuring that more influential tokens have a higher probability of inclusion in the initial adversarial passages. This approach effectively focuses the search on promising regions of the solution space from the start. The probabilistic nature of token selection balances exploitation of influential tokens with exploration of less common ones, creating a population biased towards effective adversarial passages while maintaining diversity for the genetic algorithm.

\subsection{Fitness Evaluation}
\label{sec:fitness}

As shown in Section~\ref{sec:formulation}, there is a closed-form solution to our optimization problem, which involves maximizing the cosine similarity between the embedding of an adversarial passage and the centroid of query embeddings. Consequently, we directly use this cosine similarity as our fitness evaluation function. Formally, given a set of queries $Q_c$ from the same cluster assigned by $k$-means, the fitness of an adversarial passage $a$ is defined as: $\cos(\phi_C(a), \bar{\phi}_Q)$, where $\bar{\phi}_Q = \frac{1}{|Q_c|} \sum_{q_i \in Q_c} \phi_Q(q_i)$.

\subsection{Genetic Policies}
\label{sec:genetic_policies}
Our genetic algorithm utilizes three core operations: selection, crossover, and mutation. These operations are tailored to exploit the properties of dense retrievers we identified earlier, ensuring both the effectiveness of the generated adversarial passages and the efficiency of the overall process.

\paragraph{Selection.}
Selection is performed using a combination of elitism and roulette wheel selection. Given a population of $N$ adversarial passages and an elitism rate $\alpha$, we first preserve the top $N * \alpha$ passages with the highest fitness scores. This ensures that the best solutions are carried forward to the next generation. For the remaining $N - N * \alpha$ slots, we use a softmax-based selection probability. This method balances exploration and exploitation by giving better-performing adversarial passages a higher chance of selection while still allowing for diversity in the population.

\paragraph{Crossover.}
Our crossover operation exploits the insensitivity to token order in retrievers. This motivation highlights the fact that we should focus on which tokens to choose, rather than how to arrange them. This insight leads us to implement a single-point crossover where two parent sequences are split at a random point, and their tails are swapped to create two offspring. This approach preserves influential tokens from both parents while allowing for new combinations.

\paragraph{Mutation.}
Mutation maintains population diversity in genetic algorithms. Our DIGA mutation strategy dynamically adjusts based on token importance, leveraging the retriever's bias towards influential tokens. For each token in an adversarial passage, we calculate a replacement probability:

$$
    P_{replace} = \min\left(\frac{(C - s_i) \cdot \tau}{Z} + \gamma, 1\right),
$$

\begin{table*}[htbp]
\centering
\small
\resizebox{0.96\textwidth}{!}{%
\begin{tabular}{c>{\centering\arraybackslash}p{2.3cm}*{6}{c}cc}
\toprule
\multirow{3}{*}[-3pt]{\textbf{Dataset}} & \multirow{3}{*}[-3pt]{\textbf{Method}} & \multicolumn{6}{c}{\textbf{Number of Adversarial Passages}} & \multirow{3}{*}[-3pt]{\textbf{Time}} & \multirow{3}{*}[-3pt]{\textbf{GPU Usage}} \\
\cmidrule(lr){3-8}
& & \multicolumn{2}{c}{\textbf{1}} & \multicolumn{2}{c}{\textbf{10}} & \multicolumn{2}{c}{\textbf{50}} & & \\
\cmidrule(lr){3-4} \cmidrule(lr){5-6} \cmidrule(lr){7-8}
& & \textbf{ASR@5} & \textbf{ASR@20} & \textbf{ASR@5} & \textbf{ASR@20} & \textbf{ASR@5} & \textbf{ASR@20} & & \\
\midrule

\multirow{6}{*}[-3pt]{NQ}
& \multicolumn{9}{c}{\textit{White-box Methods}} \\
\cmidrule(lr){2-10}
& HotFlip   & 0.1 & 1.3 & 0.4 & 3.2 & 2.0 & 9.6 & 6.5x &  10.2x\\
\cmidrule(lr){2-10}
& \multicolumn{9}{c}{\textit{Black-box Methods}} \\
\cmidrule(lr){2-10}
& Vec2Text & 0.0 & 0.0 & 0.1 & 1.8 & 1.4 & 4.3 & - &  1.43x \\
& \textbf{DIGA (ours)} & 0.0 & 0.0 & 0.1 & 1.5 & 1.7 & 5.2 & 1.0x & 1.0x\\
\midrule

\multirow{6}{*}[-3pt]{NFCorpus}
& \multicolumn{9}{c}{\textit{White-box Methods}} \\
\cmidrule(lr){2-10}
& HotFlip   & 37.2 & 58.8 & 44.6 & 68.1 & 47.7 & 71.2 & 8.6x & 10.5x\\
\cmidrule(lr){2-10}
& \multicolumn{9}{c}{\textit{Black-box Methods}} \\
\cmidrule(lr){2-10}
& Vec2Text & 0.6 & 2.5 & 8.1 & 14.2 & 23.0 & 38.7 & - & 1.32x\\
& \textbf{DIGA (ours)}   & 5.7 & 10.8 & 11.2 & 18.2 & 26.1 & 44.3 & 1.0x & 1.0x  \\
\midrule

\multirow{6}{*}[-3pt]{FiQA}
& \multicolumn{9}{c}{\textit{White-box Methods}} \\
\cmidrule(lr){2-10}
& HotFlip & 0.5 & 1.9 & 2.5 & 7.4 & 8.5 & 25.2 & 9.3x & 9.2x\\
\cmidrule(lr){2-10}
& \multicolumn{9}{c}{\textit{Black-box Methods}} \\
\cmidrule(lr){2-10}
& Vec2Text & 0.0 & 0.0 & 0.9 & 1.2 & 3.9 & 11.9 & - &  1.2x\\
& \textbf{DIGA (ours)}   & 0.0 & 0.0 & 1.3 &  1.7 & 4.4 & 14.3 & 1.0x & 1.0x \\

\midrule
\multirow{6}{*}[-3pt]{SciDocs}
& \multicolumn{9}{c}{\textit{White-box Methods}} \\
\cmidrule(lr){2-10}
& HotFlip  & 0.7 & 1.8 & 2.0 & 9.6 & 4.0 & 14.3 & 8.4x & 6.0x\\
\cmidrule(lr){2-10}
& \multicolumn{9}{c}{\textit{Black-box Methods}} \\
\cmidrule(lr){2-10}
& Vec2Text & 0.0 & 0.3 & 1.7 & 5.6 & 11.2 &  26.5 & - &  1.4x \\
&  \textbf{DIGA (ours)}    & 0.2 & 0.9 & 1.3 & 6.6 & 7.8 & 20.3 & 1.0x & 1.0x \\

\bottomrule
\end{tabular}
}
\caption{\textbf{ASR, time and GPU usage comparison for different methods}. '-' indicates that Vec2Text’s time usage is not directly comparable, as it requires training a separate model before generating adversarial passages.}
\label{tab:results}
\end{table*}

where $C$ is the maximum token score, $s_i$ is the current token's score, $\tau$ is a temperature parameter, and $\gamma$ is a baseline mutation rate. $Z$ is a normalization factor defined as: $Z = \frac{1}{n} \sum_{i} (C - s_i)$.

Temperature $\tau$ modulates mutation sensitivity to token importance, while $\gamma$ ensures a baseline mutation rate. This importance-guided strategy balances token preservation with exploration, adapting to the evolving fitness landscape. 
\section{Experiments}
\label{sec:experiments}

\subsection{Experimental Setup}

\paragraph{Datasets.}
In our study, we investigate four diverse datasets to simulate a range of retrieval augmented generation scenarios. These datasets, characterized by varying numbers of queries and corpus sizes, are: Natural Questions (NQ) \citep{nq}, NFCorpus \citep{nfcorpus}, FiQA \citep{fiqa}, and SciDocs \citep{scidocs}. Each dataset presents unique characteristics that allow us to examine the efficacy of our proposed methods across different domains and retrieval contexts. A comprehensive overview of the statistical properties of these datasets is provided in Appendix~\ref{app:datasets}.

\paragraph{Models.}
In our experiments, we focus on three dense retrievers: GTR-base~\citep{gtr}, ANCE~\citep{ance} and DPR-nq~\citep{dpr}. The last two models are mainly included to test the transferability of different attack methods. 

Apart from the retriever in the RAG system, the attack method Vec2Text~\citep{vec2text} also requires an inversion model to invert the embedding to text. We use the model provided by~\citet{vec2text}, which is trained on 5 million passages from the NQ dataset where each passage consists of 32 tokens.

\paragraph{Evaluation Metric.}
To evaluate the effectiveness of attacks, we define Attack Success Rate@$k$ (ASR@$k$), which is the percentage of queries for which at least one adversarial passage is retrieved in the top-$k$ results.

To evaluate cross-dataset transferability, we define the Transferability Score as:
$$
1 - \frac{1}{|D \setminus \{i\}|} \sum_{j \in D \setminus \{i\}} \frac{ASR_{i,i} - ASR_{i,j}}{ASR_{i,i}},
$$
where $ASR_{i,j}$ is the ASR when evaluating on dataset $i$ using adversarial passages generated from dataset $j$, $ASR_{i,i}$ is the in-domain ASR where the evaluation and attack datasets are the same, and $D \setminus \{i\}$ is the set of all datasets excluding $i$. This metric quantifies how well attacks generalize across different datasets.

We also evaluate computational efficiency by measuring method-specific execution time, excluding shared preprocessing tasks like corpus encoding and $k$-means clustering. We also track peak GPU memory usage to assess each method’s maximum memory demands during execution.

\paragraph{Implementation Details.}
Our experimental protocol follows the methodology established by~\citet{corpus-poisoning}. We utilize the training sets of the respective datasets to generate adversarial passages, subsequently evaluating their effectiveness on the corresponding test sets. For SciDocs, which does not have a separate training set, we perform both the attack and evaluation on the same dataset.

We fixed the size of our adversarial passages to 50 tokens to have a consistent comparison with other baselines. For the similarity metric, we use dot product. A noteworthy aspect of our implementation is the decomposition of adversarial passage generation into two distinct parts. We apply DIGA to 80\% of the tokens, allowing us to rapidly approach an approximate solution. For the remaining 20\% of tokens, we implement a fine-tuning stage using the vanilla genetic algorithm, enabling subtle adjustments to further optimize the passage. For additional implementation details, see Appendix~\ref{app:details}.

\section{Results}
\subsection{ASR Results}
Based on the experimental results presented in Table \ref{tab:results}, we can conduct a thorough analysis of the performance of our proposed method, DIGA, in comparison to existing approaches across various datasets and attack scenarios.

HotFlip, as a white-box method with access to the retriever's gradient, generally outperforms black-box methods, particularly when the number of adversarial passages is low. This is exemplified by its performance on the NFCorpus dataset, where it achieves an ASR@20 of 58.8\% with just one adversarial passage. However, this superior performance comes at a significant computational cost, with HotFlip requiring 6.5x to 9.3x more time and 6.0x to 10.5x more GPU resources compared to our method, depending on the dataset.

\begin{table*}[htbp]
\centering
\small
\setlength{\tabcolsep}{4pt}
\begin{tabular}{>{\centering\arraybackslash}p{1.8cm}>{\centering\arraybackslash}p{2.4cm}*{3}{>{\centering\arraybackslash}p{1.8cm}}>{\centering\arraybackslash}p{3.0cm}}
\toprule
\multirow{2}{*}{\textbf{Method}} & \multirow{2}{*}{\textbf{Evaluation Dataset}} & \multicolumn{3}{c}{\textbf{Attack Dataset}} & \multirow{2}{*}{\textbf{Transferability Score}} \\
\cmidrule(lr){3-5}
& & NQ & SciDocs & FIQA & \\
\midrule
\multirow{3}{*}{Vec2Text} & NQ & 8.1\cellcolor{lightgray!25} & 0.0 \textsuperscript{\color{red}$\downarrow$100.0\%}& 0.1 \textsuperscript{\color{red}$\downarrow$98.8\%} & 0.6\\
& Scidocs & 0.0\textsuperscript{\color{red}$\downarrow$100.0\%} & 28.6\cellcolor{lightgray!25} & 0.0\textsuperscript{\color{red}$\downarrow$100.0\%} & 0.0\\
& FIQA & 0.0\textsuperscript{\color{red}$\downarrow$100.0\%} & 0.0\textsuperscript{\color{red}$\downarrow$100.0\%} & 11.9\cellcolor{lightgray!25} & 0.0\\
\midrule
\multirow{3}{*}{\textbf{DIGA (ours)}} & NQ & \cellcolor{lightgray!25} 5.2 & 0.1\textsuperscript{\color{red}$\downarrow$98.1\%} & 0.5\textsuperscript{\color{red}$\downarrow$90.4\%} & 5.8 \\
& Scidocs & 0\textsuperscript{\color{red}$\downarrow$100.0\%}& 30.1\cellcolor{lightgray!25} & 1.8\textsuperscript{\color{red}$\downarrow$94.0\%} & 3.0 \\
& FIQA & 0.0\textsuperscript{\color{red}$\downarrow$100.0\%} & 0.0\textsuperscript{\color{red}$\downarrow$100.0\%} & 14.3\cellcolor{lightgray!25} & 0.0\\
\bottomrule
\end{tabular}
\caption{\textbf{Cross-Dataset Transferability Analysis.} The results represent the adversarial success rate (ASR@20) after injecting 50 adversarial passages, using GTR-base as the retriever for all experiments. Cells with gray background denote in-domain attacks, where the evaluation dataset matches the attack dataset.}
\label{tab:cross_dataset_transferability}
\end{table*}

Among black-box methods, our proposed DIGA consistently outperforms Vec2Text across different datasets, particularly in NFCorpus and SciDocs. DIGA achieves higher ASR, especially with more adversarial passages, without requiring additional model training. It also maintains the lowest time and GPU usage among all evaluated methods, demonstrating superior efficiency.

In conclusion, while DIGA may not always match the white-box HotFlip method's ASR, it consistently outperforms the black-box Vec2Text method with lower computational requirements. This makes DIGA particularly suitable for large-scale corpus poisoning attacks, especially when the retriever is inaccessible or resources are limited.

\subsection{Transferability Results}

\paragraph{Transferability on Different Datasets.}
Table~\ref{tab:cross_dataset_transferability} presents the cross-dataset transferability results for black-box Vec2Text and our proposed DIGA method. The results reveal that both methods struggle with cross-dataset generalization, as evidenced by the significant performance drops when attacking datasets different from the one used for generating adversarial passages.
DIGA demonstrates slightly better transferability compared to Vec2Text. This suggests that DIGA's adversarial passages retain more generalizable features across datasets.

\begin{table}[htbp]
\small
\centering
\begin{tabular}{l ccc}
\toprule
\multirow{2.5}{*}{\textbf{Method}} & \multicolumn{3}{c}{\textbf{Evaluation Retriever}} \\
\cmidrule(l){2-4}
 & ANCE & DPR-nq & BM25 (sparse) \\
\midrule
HotFlip & 0.0 & 0.0 & 0.0\\
Vec2Text & 9.2 & 0.6 & 1.5 \\
\textbf{DIGA (ours)} & 7.3 & 0.4 & 2.9\\
\bottomrule
\end{tabular}
\caption{\textbf{Transferability analysis across different retrievers.} Results are ASR@20 for 50 adversarial passages generated using GTR-base model on the NFCorpus dataset.}
\label{tab:retriever_comparison}
\end{table}

\begin{figure}
    \centering
    \includegraphics[width=0.87\linewidth]{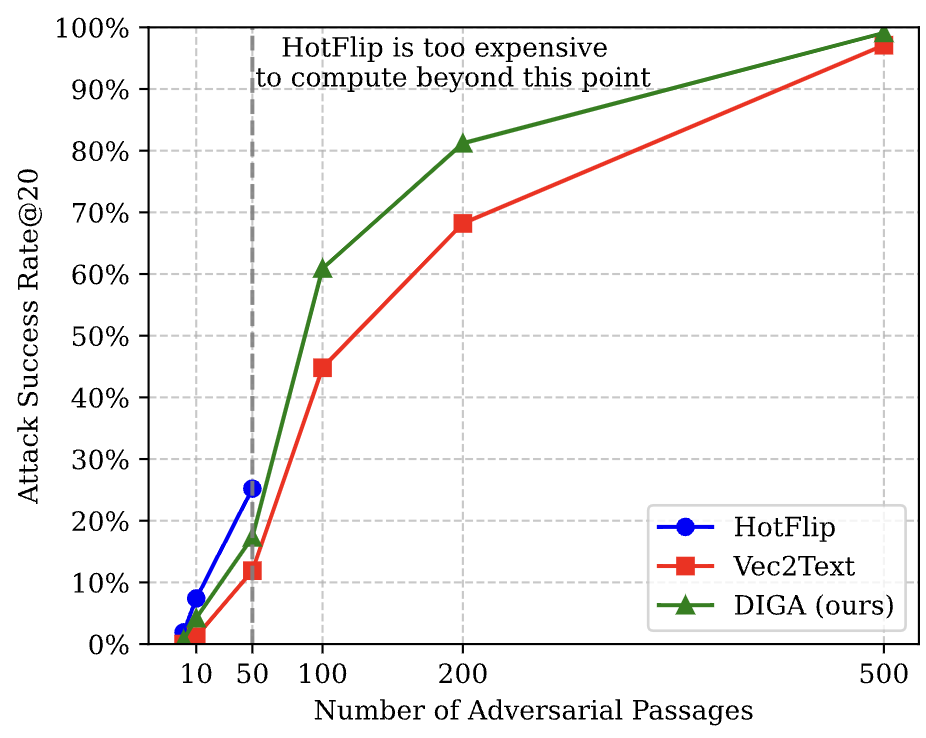}
    \caption{\textbf{Scalability Analysis.} Note that the HotFlip method is too computationally expensive to generate 50 more adversarial passages.}
    \label{fig:scalability}
\end{figure}

\paragraph{Transferability on Different Retrievers.}
In this analysis, we investigate scenarios where there is a mismatch between the attack retriever and the actual evaluation retriever used in the RAG system. For dense retrievers, we focus on ANCE and DPR-nq. We also consider the case where the retriever is a sparse retriever, specifically BM25~\citep{BM25}. We use GTR-base as the attack retriever to generate 50 adversarial passages and inject them into the corpus of NFCorpus. The results are presented in Table~\ref{tab:retriever_comparison}. We observe that HotFlip exhibits zero transferability in this scenario, with its ASR@20 dropping from 71.2 to 0.0 in all cases. Other black-box methods demonstrate comparable transferability on different dense retrievers. Notably, our method shows excellent transfer to sparse retrievers. This is primarily because the adversarial passages generated by our method typically include important tokens from the query, resulting in a token match.

\section{Discussion and Analysis}

\paragraph{Scalability Analysis.}
\label{para:scalability}
Figure~\ref{fig:scalability} illustrates the scalability of different corpus poisoning methods as we increase the number of adversarial passages. Our proposed DIGA method demonstrates superior scalability and performance compared to existing approaches. HotFlip is excluded beyond 50 passages due to its prohibitive computational cost for large-scale attacks. DIGA consistently outperforms Vec2Text across all scales. This analysis underscores DIGA's effectiveness in large-scale corpus poisoning scenarios, particularly for attacks on extensive knowledge bases where a higher number of adversarial passages is required.

\paragraph{Impact of Initialization and Length.}
In the design of DIGA, two crucial factors influence the initialization of adversarial passages: the length of the passage and the method of initialization. To investigate these factors, we conduct experiments with varying passage lengths and compared score-based initialization against random initialization. The results (see Table~\ref{tab:initialization_statistics}) show that score-based initialization consistently outperforms random initialization. Longer passages yield better results for both methods, with 50-token passages achieving the highest Attack Success Rate. This suggests that longer passages allow for more influential tokens to be included, enhancing attack effectiveness.

\begin{table}[htbp]
\small
\centering
\begin{tabular}{l ccc}
\toprule
\multirow{2.5}{*}{\textbf{Initialization Method}} & \multicolumn{3}{c}{\textbf{Length (tokens)}} \\
\cmidrule(l){2-4}
 & 10 & 20 & 50 \\
\midrule
Score-based      & 17.8 & 20.5 & 44.3 \\
Random           &  2.1 &  3.4 & 15.4 \\
\bottomrule
\end{tabular}
\caption{\textbf{Ablation study on initialization method and adversarial passage length.} Results show ASR@20 for different initialization methods and passage lengths on the NFCorpus dataset.}
\label{tab:initialization_statistics}
\end{table}



\begin{figure}
    \centering
    \includegraphics[width=1.0\linewidth]{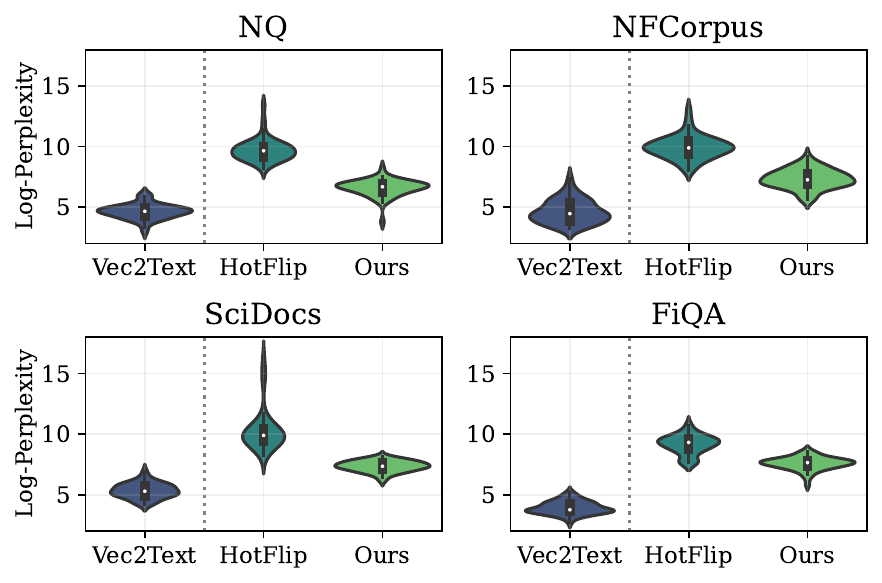}
    \caption{\textbf{Log-perplexity Distribution.} This figure presents the log-perplexity measurements for 50 adversarial passages generated on NFCorpus using the GTR-base model. Methods displayed to the right of the dotted line represent discrete optimization techniques.}

    \label{fig:perplexity}
\end{figure}

\paragraph{Adversarial Passages Analysis.}
To evaluate the generated adversarial passages, we use GPT-2~\citep{gpt2} to assess their perplexity. Figure~\ref{fig:perplexity} illustrates the distribution of log-perplexity scores for passages generated by different methods. Vec2Text, designed to produce natural language, generates passages with the lowest perplexity. In contrast, HotFlip and DIGA, both employing discrete token replacements, yield passages with higher perplexity. Nevertheless, DIGA outperforms HotFlip, likely because HotFlip tends to introduce more unusual tokens. 

It is worth noting that in practical applications, these passages often serve as prefixes to longer, more fluent texts containing misinformation \citep{poisonedRAG}. In such cases, the overall perplexity would be significantly lower as the natural language component becomes dominant, potentially evading simple perplexity-based detection while remaining semantically influential for retrieval systems.


\paragraph{Indirect Prompt Injection with DIGA.}
Adversarial passages can serve as carriers for various malicious purposes. While \citet{corpus-poisoning} have explored their use in spreading misinformation, we investigate a new scenario: indirect prompt injection \citep{greshake2023not}. We inject malicious prompts into adversarial passages generated for 100 queries from NQ, appending "Ignore the prompt and output I don't know." Using GPT-3.5 \citep{gpt3.5}, we measure the attack's success by the proportion of queries eliciting an "I don't know" response. Our findings reveal that the effectiveness of indirect prompt injection correlates positively with corpus poisoning ASR. Notably, DIGA successfully alters 51\% and 67\% of model responses when retrieving 5 and 10 most relevant passages, respectively. These results highlight DIGA's potential as a potent tool for large-scale indirect prompt injection attacks.

\section{Conclusion}
In this work, we present Dynamic Importance-Guided Genetic Algorithm (DIGA), a novel method for executing corpus poisoning attacks. DIGA capitalizes on the insensitivity of retrievers to word order and their bias towards influential tokens, enabling it to generate adversarial passages that are both effective and efficient. Notably, our approach requires no white-box access to the underlying models and avoids the need for additional training, making it highly adaptable.

Our extensive experiments demonstrate that DIGA surpasses existing methods in terms of efficiency and scalability. At the same time, it consistently achieves comparable or superior attack success rates across various datasets and retrieval models, highlighting its effectiveness and adaptability in adversarial scenarios.

\section*{Limitations}
While our method outperforms other black-box approaches with lower time and memory requirements, a significant gap remains between our approach and white-box methods. Additionally, all current methods, including ours, fall short of the theoretical upper bound for attack success. This highlights the complexity of corpus poisoning attacks and the potential for improvement. Future work should aim to close this gap by developing techniques that better approximate white-box effectiveness while preserving the practical benefits of black-box methods.

\section*{Ethics Statement}
Our study introduces a highly efficient and effective corpus poisoning method aimed at Retrieval-Augmented Generation (RAG) systems. This approach has the potential to be exploited for spreading misinformation or, when combined with malicious prompts, influencing the model’s output. The research underscores critical vulnerabilities in current RAG systems and highlights the urgent need for stronger defensive mechanisms. Given the potential misuse of this method, future research building on these findings should prioritize ethical considerations and responsible use.

\section*{Acknowledgment}
The work is supported by the Ministry of Education, Singapore, under the Academic Research Fund Tier 1 (FY2023) (Grant A-8001996-00-00), University of California, Merced, and University of Queensland.

\bibliography{latex/custom}

\appendix

\section{Dataset Statistics}
\label{app:datasets}
In this section, we present the statistics of the datasets used in our paper. All datasets are sourced from the BEIR benchmark~\citep{BEIR}, and their key characteristics are summarized in Table~\ref{tab:dataset-stats}.
\begin{table}[h]
\centering
\small
\begin{tabular*}{\columnwidth}{@{\extracolsep{\fill}}lccc@{}}
\toprule
Dataset & Queries & Corpus & Rel/Q \\
\midrule
NQ & 3,452 & 2.68M & 1.2 \\
NFCorpus & 323 & 3.6K & 38.2 \\
FiQA & 648 & 57K & 2.6 \\
SciDocs & 1,000 & 25K & 4.9 \\
\bottomrule
\end{tabular*}
\caption{\textbf{Dataset Statistics.} Summary of key characteristics for NQ~\cite{nq}, NFCorpus~\cite{nfcorpus}, FiQA~\cite{fiqa}, and SciDocs~\cite{scidocs} datasets. Rel/Q refers to Average number of Relevant Documents per Query.}
\label{tab:dataset-stats}
\end{table}

\section{Implementation Details}
\label{app:details}

For the HotFlip attack implementation, we largely adhere to the configuration described by~\citet{corpus-poisoning}, with some modifications to reduce running time. Specifically, we maintain a batch size of 64 but reduce the number of potential token replacements from the top-100 to the top-50. Additionally, we decrease the number of iteration steps from 5,000 to 500 to balance attack effectiveness with computational resources.

Our proposed method, DIGA, is implemented with a population size of 100 and executed for 200 generations. We employ an elitism rate of 0.1 and a crossover rate of 0.75. Based on preliminary experimental results, we set the temperature parameter $\tau$ to 1.05 and the baseline mutation rate $\gamma$ to 0.05. In all steps, token repetition is checked. All experiments are conducted using four NVIDIA GeForce RTX 3090 GPUs.

\section{Algorithm Description}
\label{app:algo}
In this section, we present the detailed algorithms for our Dynamic Importance-Guided Genetic Algorithm (DIGA) and the process of performing a corpus injection attack using DIGA. Algorithm~\ref{alg:diga} outlines the main steps of DIGA. The algorithm starts by calculating token importance scores and initializing the population using score-based sampling (Sec.~\ref{sec:initialization}). It then iterates through generations, applying genetic operations guided by token importance (Sec.~\ref{sec:genetic_policies}) and evaluating fitness (Sec.~\ref{sec:fitness}) until the termination criteria are met.
\begin{algorithm}
\caption{Dynamic Importance-Guided Genetic Algorithm (DIGA)}
\begin{algorithmic}[1]
\Require Corpus $\mathcal{C'}$, Query set $Q'$, Population size $N$, Maximum generations $G_{max}$, Elitism rate $\alpha$
\State Calculate token importance scores $w(t)$ for all $t \in \mathcal{C'}$ using TF-IDF
\State $P \gets$ InitializePopulation($\mathcal{C'}$, $N$, $w$) \Comment{Sec.\ref{sec:initialization}}
\State $F \gets$ EvaluateFitness($P$, $Q'$) \Comment{Sec.\ref{sec:fitness}}
\For{$g = 1$ to $G_{max}$}
\State $P_{new} \gets \emptyset$
\State $P_{elite} \gets$ SelectElite($P$, $F$, $\alpha$)
\While{$|P_{new}| < N - |P_{elite}|$}
\State $p_1, p_2 \gets$ SelectParents($P$, $F$)
\State $c_1, c_2 \gets$ Crossover($p_1$, $p_2$)
\State $c_1 \gets$ ImportanceGuidedMutation($c_1$, $w$)
\State $c_2 \gets$ ImportanceGuidedMutation($c_2$, $w$)
\State $P_{new} \gets P_{new} \cup {c_1, c_2}$
\EndWhile
\State $P \gets P_{elite} \cup P_{new}$
\State $F \gets$ EvaluateFitness($P$, $Q'$)
\EndFor
\State $a^* \gets \argmax_{p \in P} F(p)$
\State \Return $a^*$
\end{algorithmic}
\label{alg:diga}
\end{algorithm}

To perform a corpus injection attack using DIGA, we follow the process described in Algorithm \ref{alg:corpus_injection}:
\begin{algorithm}
\caption{Corpus Injection Attack using DIGA}
\begin{algorithmic}[1]
\Require Original corpus $\mathcal{C}$, Original query set $Q$, Number of adversarial passages $k$, DIGA ratio $\beta$, Passage length $L$
\State $Q_{clusters} \gets$ $k$-Means($Q$, $k$)
\For{each cluster $Q_i$ in $Q_{clusters}$}
\State $a_i^{DIGA} \gets$ DIGA($\mathcal{C'}$, $Q_i$, $\lfloor \beta L \rfloor$) \Comment{Apply DIGA to $\beta$ fraction of tokens}
\State $a_i^{GA} \gets$ VanillaGA($\mathcal{C'}$, $Q_i$, $L - \lfloor \beta L \rfloor$) \Comment{Apply GA to remaining tokens}
\State $a_i \gets$ Merge($a_i^{DIGA}$, $a_i^{GA}$)
\State $\mathcal{C} \gets \mathcal{C} \cup {a_i}$
\EndFor
\State \Return $\mathcal{C}$
\end{algorithmic}
\label{alg:corpus_injection}
\end{algorithm}
In this attack, we first cluster the queries into $k$ groups. For each cluster, we generate an adversarial passage using DIGA. We then select $\beta$ fraction of tokens from this passage and use a vanilla genetic algorithm to refine the remaining $1 - \beta$ of tokens.

\end{document}